\documentclass[10pt,twocolumn,letterpaper]{article}

\usepackage{cvpr}              
\usepackage{multirow}
\usepackage[normalem]{ulem}
\useunder{\uline}{\ul}{}
\usepackage{booktabs}
\usepackage{graphicx}

\definecolor{cvprblue}{rgb}{0.21,0.49,0.74}
\usepackage[pagebackref,breaklinks,colorlinks,allcolors=cvprblue]{hyperref}

\def\paperID{} 
\def\confName{CVPR}
\def\confYear{2026}

\title{InnoAds-Composer: Efficient Condition Composition for E-Commerce Poster Generation}

\author{
Yuxin Qin\textsuperscript{1*}, Ke Cao\textsuperscript{1*}, Haowei Liu\textsuperscript{2*}, Ao Ma\textsuperscript{1\dag}, Fengheng Li\textsuperscript{1}, Honghe Zhu\textsuperscript{1}, \\ Zheng Zhang\textsuperscript{1\ddag}, 
Run Ling\textsuperscript{1}, Wei Feng\textsuperscript{1}, Xuanhua He\textsuperscript{3}, Zhanjie Zhang\textsuperscript{4\ddag}, Zhen Guo\textsuperscript{1}, Haoyi Bian\textsuperscript{1}, \\ Jingjing Lv\textsuperscript{1}, Junjie Shen\textsuperscript{1}, 
Ching Law\textsuperscript{1} \\
\textsuperscript{1}JD.com, Inc., Beijing, China \\
\textsuperscript{2}Chongqing University of Posts and Telecommunications, Chongqing, China \\
\textsuperscript{3}The Hong Kong University of Science and Technology, Hong Kong, China \\
\textsuperscript{4}Zhejiang University, Hangzhou, China \\
{\tt\small maao.8@jd.com, zhangzhanj@126.com}
}

\begin{document}

\twocolumn[{
\renewcommand\twocolumn[1][]{#1}
\maketitle
\begin{center}
    \centering
    \vspace*{-.4cm}
    \includegraphics[width=\textwidth]{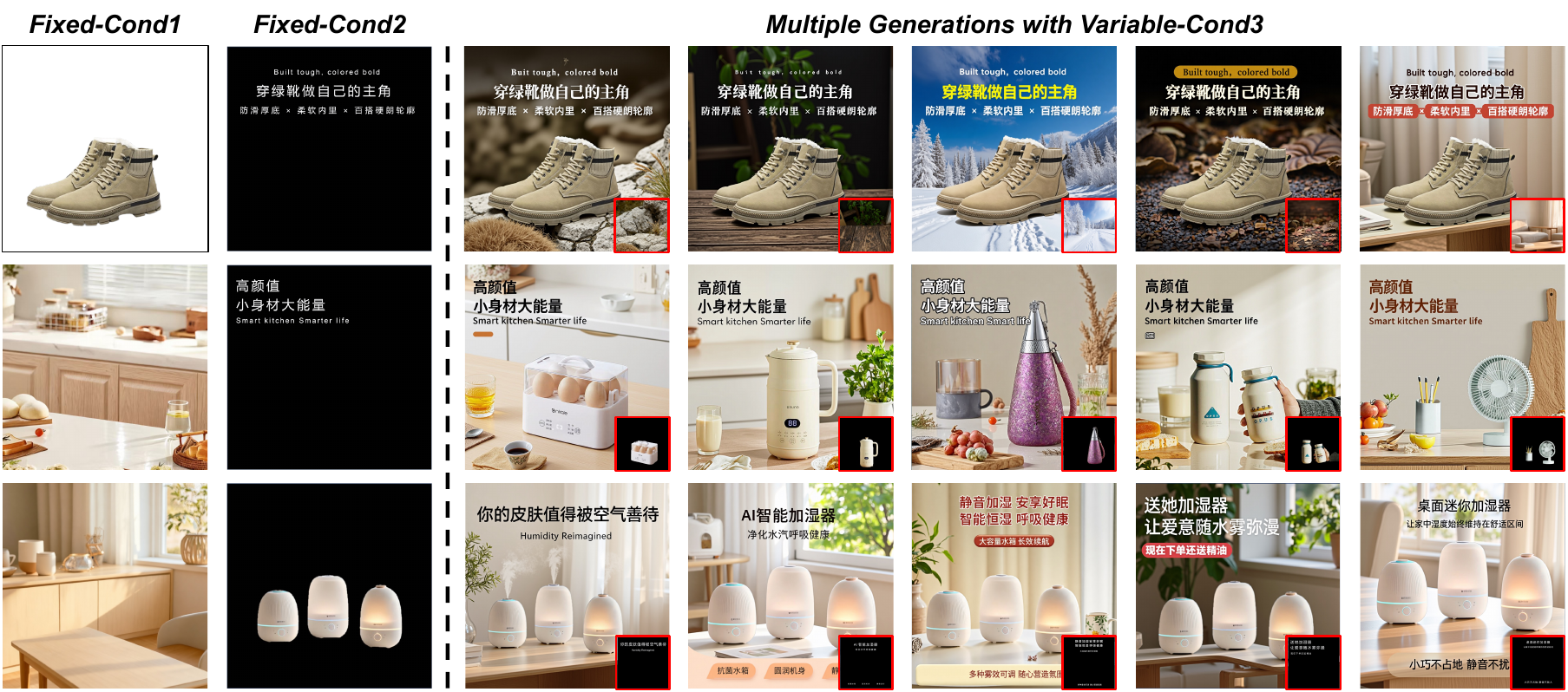}
    \vspace*{-.3cm}
    \captionof{figure}{Inno-Composer generates high-quality e-commerce posters under three independent controls—background style, subject appearance, and glyph text. Each row varies a single condition while keeping the other two fixed. The bottom-right inset shows the input of the varied condition.}
\label{fig:fig1}
\end{center}
}]

\begingroup
\renewcommand\thefootnote{}%
\footnotetext{%
\textsuperscript{*} Equal contribution.\par
\hspace{\parindent}\textsuperscript{\dag} Project leader.\par
\hspace{\parindent}\textsuperscript{\ddag} Corresponding author.%
}
\endgroup

\maketitle
\begin{abstract}
E-commerce product poster generation aims to automatically synthesize a single image that effectively conveys product information by presenting a subject, text, and a designed style. Recent diffusion models with fine-grained and efficient controllability have advanced product poster synthesis, yet they typically rely on multi-stage pipelines, and simultaneous control over subject, text, and style remains underexplored. Such naive multi-stage pipelines also show three issues: poor subject fidelity, inaccurate text, and inconsistent style.
To address these issues, we propose InnoAds-Composer, a single-stage framework that enables efficient tri-conditional control tokens over subject, glyph, and style. To alleviate the quadratic overhead introduced by naive tri-conditional token concatenation, we perform importance analysis over layers and timesteps and route each condition only to the most responsive positions, thereby shortening the active token sequence. Besides, to improve the accuracy of Chinese text rendering, we design a Text Feature Enhancement Module (TFEM) that integrates features from both glyph images and glyph crops.
To support training and evaluation, we also construct a high-quality e-commerce product poster dataset and benchmark, which is the first dataset that jointly contains subject, text, and style conditions. Extensive experiments demonstrate that InnoAds-Composer significantly outperforms existing product poster methods without obviously increasing inference latency.
\end{abstract}

\section{Introduction}
\label{sec:intro}
E-commerce product poster generation has emerged as a crucial task that aims to automatically synthesize a single image effectively conveying product information through the integration of subject, text, and a designed style. Recently, diffusion models~\cite{rombach2022high,chen2023pixarta,stabilityai2024sd3,flux2024,he2024id,feng2025fancyvideo,ma2025lay2story,wang2025pt,zhang2024artbank,wangwisa,liu2026hifi,
zhou2025identitystory,
zhou2024magictailor,
song2025scenedecorator,
chen2025empirical,
song2025hero} have demonstrated fine-grained and efficient control over image synthesis, achieving high visual fidelity and semantic richness, which has greatly promoted the development of automated poster design. However, e-commerce poster generation remains relatively underexplored. Unlike free-form artistic layouts, retail posters must follow strict layout and branding rules while maintaining subject fidelity, style consistency, and text accuracy, making poster creation a constrained and multi-objective problem.

Current e-commerce poster generation systems still fall short in three respects. First, most do not offer end-to-end joint control of background style, subject fidelity, and text accuracy within a single model; multi-stage pipelines~\cite{cao2024product2img,gao2023textpainter} that compose the scene and render the text tend to be inaccurate, resulting in style inconsistency and loss of subject fidelity. Second, emerging single-stage approaches~\cite{chen2023diffute,ma2025glyphdraw2} incorporate text control but struggle to render complex scripts and small glyphs with high fidelity. Third, the designed background style is often prompt-driven and may deviate from global style or semantic constraints ~\cite{gao2025postermaker,zhang2025creatidesign,zhang2025u,zhang2025dyartbank,zhang2025spast}. These limitations are exacerbated by the scarcity of training data: e-commerce poster datasets with fine-grained, multi-condition annotations are limited, hindering the learning of reliable design priors and robust controllability.

To address these challenges, we introduce InnoAds-Composer, a single-stage, multi-condition framework for e-commerce poster synthesis built on an MM-DiT backbone. A unified tokenization maps style, subject, and glyph conditions into the same token space, enabling joint inference while preserving the priors of the underlying text-to-image model. To remedy the weakness of text rendering, we propose a Text Feature Enhancement Module (TFEM): one branch encodes the entire glyph image with a VAE to obtain visual glyph tokens; a second branch processes single-glyph crops with an OCR backbone and injects multi-positional cues (absolute location, font size, and local position). For the above single/entire glyph tokens, we use a lightweight character encoder and then fuse them, improving glyph sharpness, boundary integrity, and readability in a principled way. To curb computation and avoid redundant conditioning, we conduct layer and timestep importance analysis, estimating the backbone’s importance differences for the three conditions and performing importance-aware injection. By retaining each condition only at its most responsive layers and diffusion steps, we shorten the effective sequence and temper the quadratic growth of attention. In implementation, a decoupled attention design preserves the main stream’s sensitivity to conditions while removing costly, low-value interactions, yielding consistent efficiency gains in both training and inference. To support training and evaluation, we also create a high-quality e-commerce poster dataset and benchmark. This data pipeline is designed to provide diverse background style control, along with accurate subject and glyph controls, providing the supervision needed for robust multi-condition generation. 
Qualitative results in Figure~\ref{fig:fig1} show that InnoAds-Composer produces high-quality posters under three independent controls.

Our contributions can be summarized as follows:
\begin{itemize}
\item We propose InnoAds-Composer, a single-stage framework for e-commerce posters that provides efficient, coordinated control over style, subject, and text. With TFEM, the model fuses entire glyph and single glyph features augmented by positional cues, systematically improving text accuracy.

\item We reveal non-uniform and complementary patterns of condition influence across layer and timestep, and leverage these differences to inject tokens only at the most responsive layers and steps, reducing the activated token sequence and restraining attention's quadratic complexity growth.

\item We design a new data construction pipeline and release InnoComposer-80K together with InnoComposer-Bench, covering subject, glyph, style, and overall quality, enabling unified comparison for e-commerce poster generation.

\item Extensive experiments show that our method effectively addresses the core difficulties of e-commerce poster generation, achieving strong visual quality and control while significantly lowering inference cost.

\end{itemize}

\section{Related Work}
\label{sec:formatting}
Please see Appendix Sec.~\ref{appendix:relatedworks}.
\section{Datasets and Benchmark}
\label{sec:Datasets}
\begin{figure*}[h]
    \centering
    \includegraphics[width=0.9\textwidth]{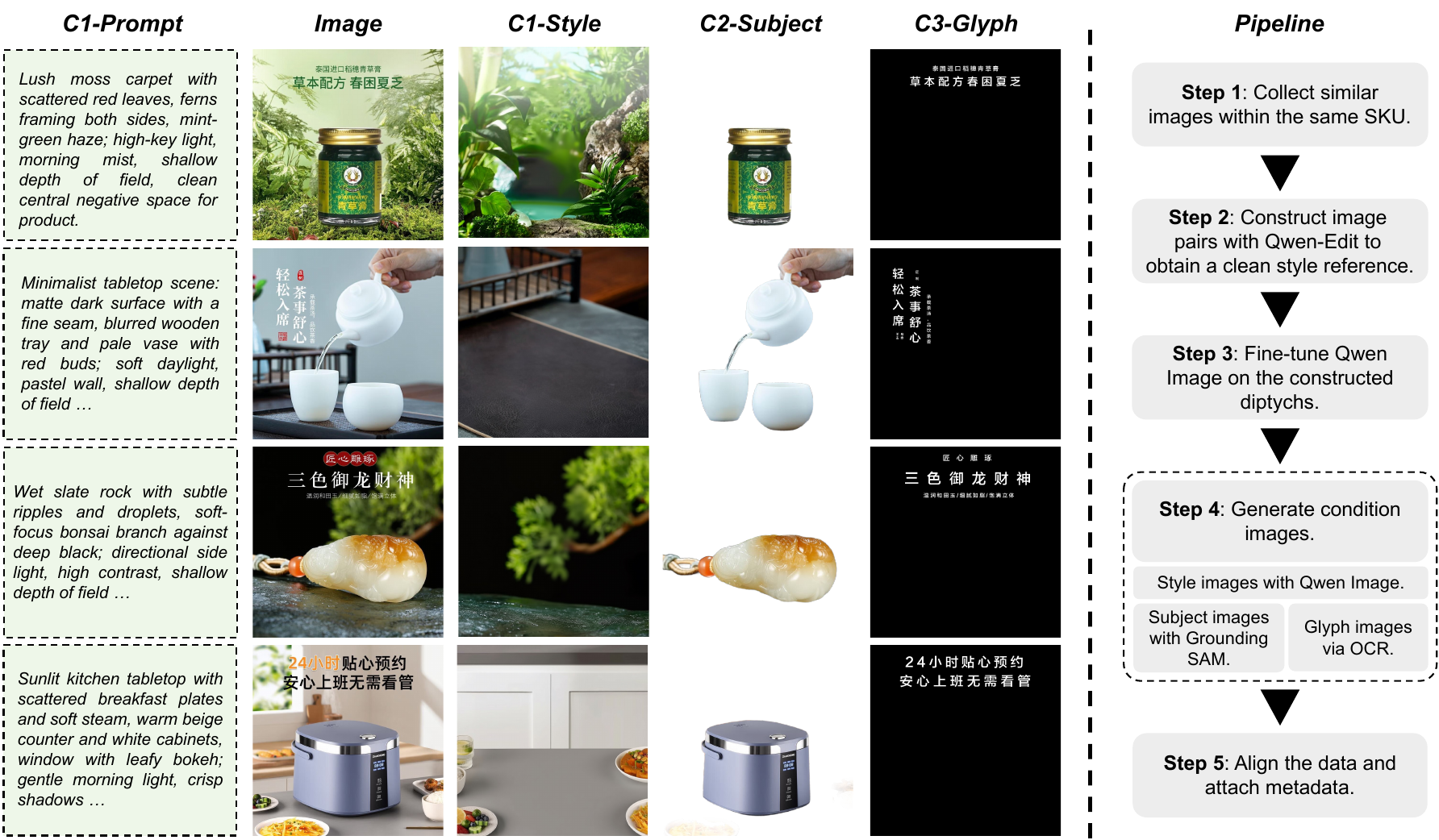}
    \caption{\label{fig-data}Case Examples and Dataset Construction Pipeline for E-commerce Poster Generation.}
    \label{fig-data}
    \vspace{-10pt}
\end{figure*}
As shown in Figure~\ref{fig-data}, we construct a high-quality bilingual e-commerce poster dataset through a structured pipeline. The left panel presents typical cases composed of three customized inputs: a background-style image that conveys global aesthetics and background elements, a subject image that specifies the product and its approximate location, and a glyph image that provides the textual content to be rendered. The right panel summarizes the pipeline. Images are first grouped by Stock Keeping Unit (SKU), a unique identifier for each distinct product, and annotators select visually similar pairs within each group. For one image in each pair, Qwen-Edit~\cite{wu2025qwenimage} removes product content to obtain a clean style reference; when needed, a super-resolution model improves clarity. The reference and its paired image are concatenated into a diptych, and the reference side is recorded. These diptychs then supervise a fine-tuning of Qwen-Image~\cite{wu2025qwenimage} so the generator learns to produce style-consistent counterparts, yielding clean background references. Using the fine-tuned generator, we synthesize style images in batches. Unlike conventional pipelines that enforce a tight pixel-level match between the content image and its style counterpart, our synthesized backgrounds are semantically aligned yet intentionally differ in local details, which fosters diversity during generation. The Glyph images containing both Chinese and English are extracted from the original images through OCR, and subject images are obtained by segmenting the product with Grounded-SAM~\cite{ren2024groundedsam,wang2025learning} to produce foregrounds or masks. Finally, we curate the data, perform de-duplication and resolution normalization, and record metadata such as SKU and preprocessing parameters. Following this pipeline, we construct InnoComposer-80K, a corpus of 80,000 poster samples. Each sample contains a text prompt, a subject image, a background-style image, and a glyph image, providing comprehensive supervision for multi-condition poster generation. For evaluation, we curate InnoComposer-Bench, a 300-item subset ranked by product emphasis, style consistency, and text accuracy; these items are strictly held out from training to ensure fair comparison across methods.

\section{Methods}
\label{sec:Methods}
\subsection{InnoAds-Composer}
\begin{figure*}[h]
    \centering
    \includegraphics[width=\textwidth]{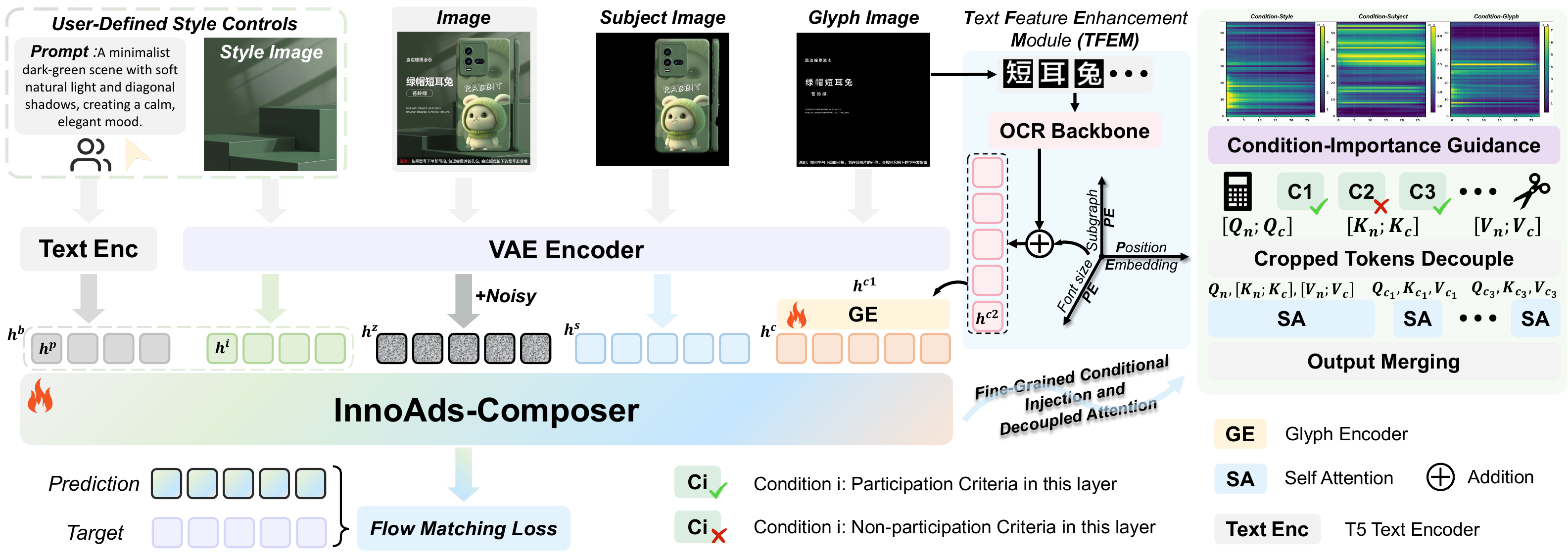}
    \caption{\label{fig-main}\textbf{Overview of InnoAds-Composer.} The framework comprises three modules: (1) \emph{Multi-Condition Tokenization}, which maps heterogeneous controls into a shared token space and aligns them with the MM-DiT backbone; (2) \emph{Importance-Aware Condition Injection}, which routes each control to its importance layers to improve efficiency while preserving controllability; and (3) \emph{Decoupled Attention}, which allows the main stream to attend to condition cues while the condition branch performs self-attention only, removing the extra path to reduce cost and maintain training–inference consistency.}
    \label{fig-main}
    \vspace{-10pt}
\end{figure*}
Figure~\ref{fig-main} illustrates InnoAds-Composer, a multi-condition, text-to-image advertising framework built on an MM-DiT backbone. A user prompt is first processed by a pretrained T5 text encoder, which tokenizes and embeds the prompt into a sequence of text tokens $h^p$. In parallel, a reference image is mapped by the VAE encoder into a latent representation $h^z$, which is then partitioned into patches to form a set of visual tokens. The text and visual tokens are unified and propagated through MM-Attention blocks that interleave intra-modal self-attention with cross-modal interactions. This design aligns language semantics with fine-grained visual structure, enabling precise control over the generated content.

To meet e-commerce requirements that require creatives to honor product focus, layout constraints, brand elements, and style specifications, InnoAds-Composer accepts multiple heterogeneous conditions, such as product attributes, layout hints, brand text or logos, and stylistic tags. These conditions are embedded in the same token space and injected across MM-DiT layers, guiding generation while preserving the strong priors of the underlying text-to-image model. Operating end-to-end in the latent space, the system maintains the image quality and diversity of the base model and delivers high-fidelity, efficiently controllable poster synthesis tailored to e-commerce use cases.

\subsubsection{Multi-Condition Tokenization}
To enable high-quality, controllable e-commerce poster generation, InnoAds-Composer adopts a unified multi-condition tokenization strategy. Heterogeneous controls, including global background style, subject imagery, and text layout, are mapped to a shared embedding space and injected across MM-DiT through MM-Attention, preserving the base T2I model’s priors while enabling precise end-to-end control.

\textbf{User-Defined Background Style Control.} 
Global style is specified either by a style prompt or by a style image, handled within a single formulation. Pretrained text encoder tokenizes the prompt into text tokens $h^p$, whereas a style image is encoded by a VAE into a latent grid and patchified into visual tokens $h^i$.The resulting background-style tokens are defined as:
\begin{equation}\label{}
h^{b} =
\begin{cases}
\mathcal{C} \left(h^{p}\right), 
& m=0\\
\mathcal{C}\left(h^{i},h^{p_0}\right), 
& m=1.
\end{cases}
\end{equation}
where $m=1$ indicates the presence of a style image, $\mathcal{C}$ represents the concatenation. $h^{p_0}$ is a fixed anchor prompt independent of user inputs;

\textbf{Subject Control.} 
To emphasize the product foreground and suppress background leakage, regions outside the subject are filled with black to form an explicit mask. The masked image is then encoded by the VAE and patchified, producing subject tokens $h^s$ that capture the object’s structure and appearance while remaining aligned to the shared token space.

\textbf{Glyph Control with Text-Feature Enhancement.} 
Readable, well-placed text is pivotal for conversion in ad creatives. We therefore introduce a dual-branch glyph control with a Text Feature Enhancement Module (TFEM). In the first branch, the entire glyph image is encoded by a VAE and patchified to produce visual glyph tokens $h^{c1}$. In the second branch, single-glyph crops extracted from the original glyph image are first processed by an OCR backbone; afterward, three positional encodings are added: the absolute position in the original image, a font-size code indicating the intended scale, and a local positional encoding within each crop, yielding $h^{c2}$. A lightweight Character Encoder then fuses both sources:
\begin{equation}\label{}
h^c=\mathbf{GlyphEnc}(h^{c1},h^{c2})
\end{equation}
yielding glyph tokens that jointly encode glyph fidelity, encompassing clarity and edge integrity, together with semantic and positional intent expressed through legibility and alignment.

\subsubsection{Importance-Aware Condition Injection}
Building on multi-condition tokenization, we estimate the MM-DiT backbone’s preferences for background, subject, and character conditions across layers and diffusion timesteps in Sec .~\ref {Conditions Importance Analysis}. The resulting preference curves indicate where each condition exerts the strongest influence. Guided by these curves and validated through ablation, we retain by default 40\% of style tokens, 50\% of subject tokens, and 20\% of glyph tokens. In any layer selected for condition type $i$, the corresponding condition tokens are concatenated with the main-stream noisy-latent tokens $h^{z}$, and all non-selected condition tokens are omitted at that layer. This targeted scheduling shortens the active sequence per layer and substantially curbs token-induced computational growth in MM-DiT, while preserving precise controllability.

\subsubsection{Decoupled Attention}
In token-based conditional diffusion processes, condition tokens generally evolve slowly across timesteps and their representations remain largely unaffected by the noisy latent tokens~\cite{he2025fulldit2,wang2022spnet,hong2026stymam,cao2024shuffle}. Applying full attention over the concatenated sequence \(\left[ \mathbf{h}_n; \mathbf{h}_c \right]\) therefore incurs redundant computation, as it repeatedly processes interactions that are either nearly static across steps or provide minimal useful signal relative to their computational cost.

We address this by removing the pathway from condition queries to noisy-latent keys while retaining the pathway from noisy-latent queries to condition keys. Conditions continue to guide generation through the mainstream, and the condition branch no longer follows rapidly changing noise. Let $\mathbf{Q}_n,\mathbf{K}_n,\mathbf{V}_n$ be the queries, keys, and values of the noisy-latent tokens ${h^z}$, and $\mathbf{Q}_{ci},\mathbf{K}_{ci},\mathbf{V}_{ci}$ those of condition type $i$. We compute
\begin{align}
  {{O}_{n}}& =\mathbf{Attn}({{Q}_{n}},\left[ {{K}_{n}};{{K}_{ci}} \right],\left[ {{V}_{n}};{{V}_{ci}} \right]) \\ 
  {{O}_{ci}}& =\mathbf{Attn}({{Q}_{c}},{{K}_{ci}},{{V}_{ci}}) \\ 
  O& =[{{O}_{n}},{{O}_{ci}}]
\end{align}
The main stream attends to both its own context and the condition cues, whereas the condition stream performs self-attention only. This eliminates the \(\mathbf{Q}_c\text{--}\mathbf{K}_n\) cross-attention term during both training and inference, reducing computation while preserving consistency. Moreover, since the condition stream no longer depends on the timestep, its activations at each block can be computed once  and cached for reuse across all timesteps during inference.


\subsubsection{Two-Stage Training Strategy}
Concatenating multiple condition tokens inflates the sequence length and, in turn, the attention cost. To control complexity, we prune tokens guided by condition-importance analysis and adopt a two-stage training. In Stage I, all condition tokens are retained to train a fully conditioned poster generator. In Stage II, we remove the selected tokens and fine-tune the network. During this phase, diffusion timesteps are sampled in proportion to their mass in the global importance map, thereby aligning the training emphasis with the importance distribution observed at evaluation. This procedure mitigates the performance drop from pruning, preserves the generative capacity of the full model, and substantially reduces inference-time computation.

\subsection{Conditions Importance Analysis}
\label{Conditions Importance Analysis}
\begin{figure}[h]
    \centering
    \includegraphics[width=\linewidth]{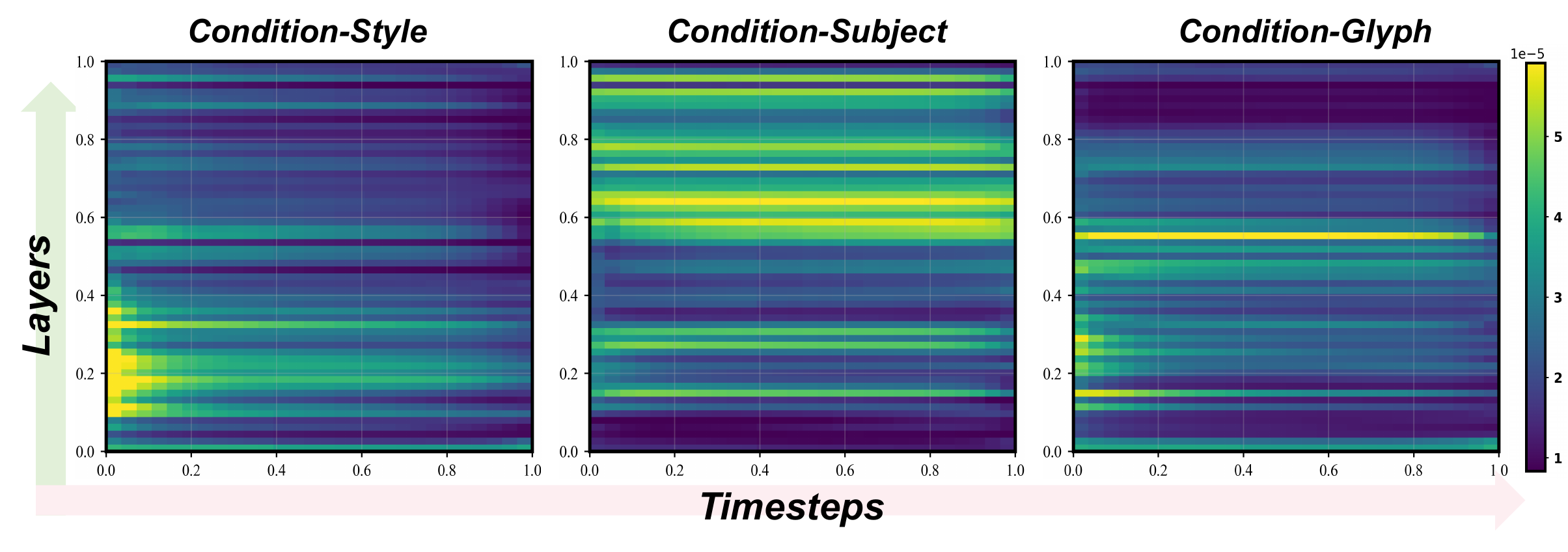}
    \vspace{-20pt}
    \caption{\label{fig-main-importance}The importance heatmaps of the three conditions across timesteps and layers.}
    \label{fig-main-importance}
    \vspace{-10pt}
\end{figure}

\begin{figure*}[t]
\centering
\includegraphics[width=0.95\linewidth]{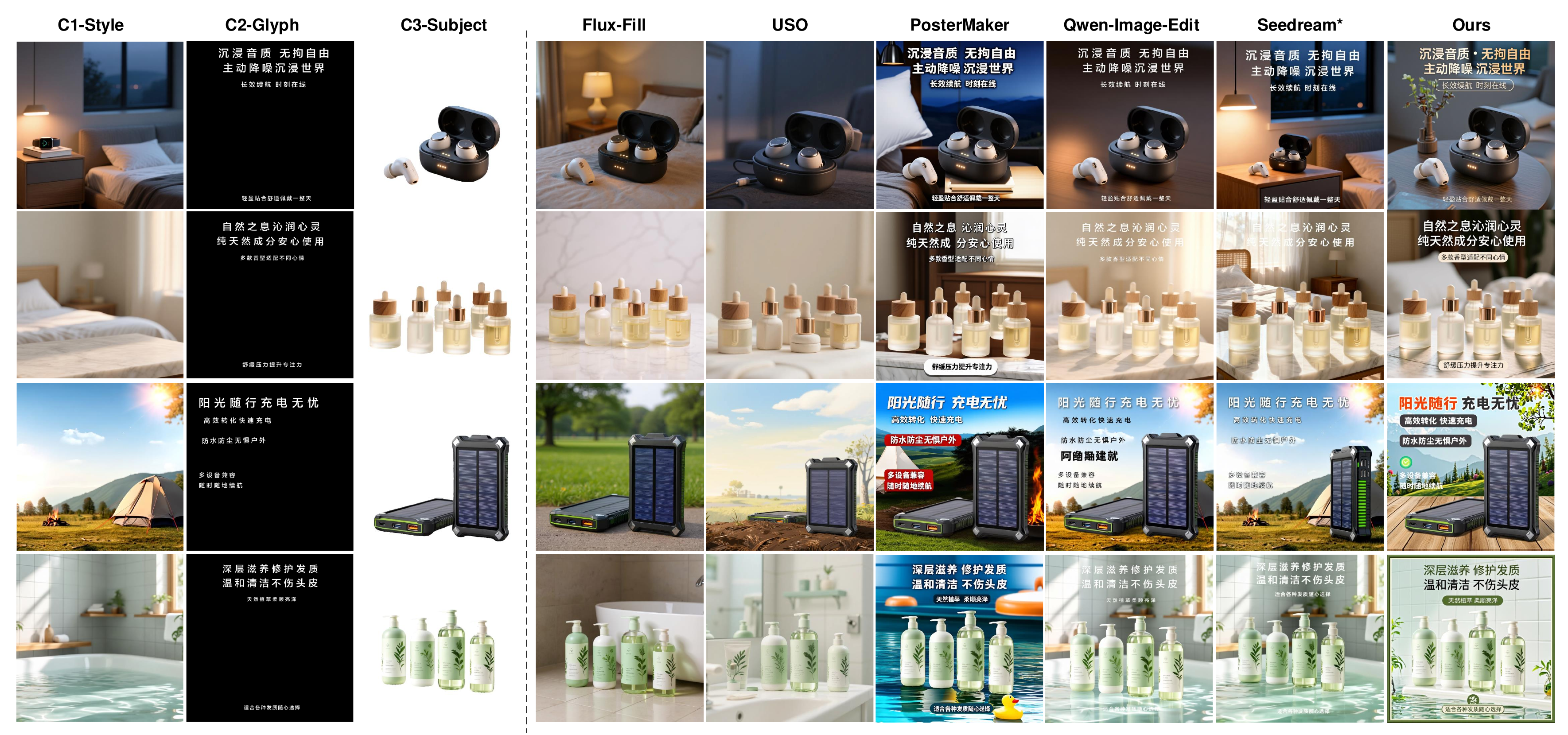}
\vspace{-10pt}
\caption{Qualitative results. \textbf{Left:} Input conditions, including C1-style images, C2-glyph images, and C3-subject images. \textbf{Right:} Results generated by different methods.}
\label{fig:comp_gen}
\vspace{-10pt}
\end{figure*}

Self-attention in transformers typically dominates both computational cost~\cite{jiang2023res,bi2024using} and memory usage~\cite{xu2025dropoutgs,cao2024frequency,jiao2025mapomotionawarepartitioning} due to its quadratic complexity~\cite{cao2025relactrl}. In MM-DiT, this issue is further amplified by the concatenation of condition tokens with noisy-latent tokens, significantly increasing the sequence length and exacerbating the computational bottleneck. Specifically, in e-commerce poster generation, the introduction of diverse conditions such as reference images, subject representations, and text content offers precise control over the overall style, product depiction, and text clarity. However, this increased flexibility comes at a high computational cost, particularly when these conditions are uniformly injected across all layers and timesteps of the model. To address this challenge, we conduct a detailed analysis of the importance of different control conditions across various layers and timesteps in MM-DiT. By investigating how each condition influences the generation process at different stages, we identify the most effective layers and timesteps for injecting each condition. This allows us to determine where each condition has the most significant impact and where it can be omitted to improve computational efficiency. Based on these insights, we propose a selective injection strategy. In this strategy, each condition is routed only to its most relevant layers and timesteps, while less responsive conditions are omitted. This selective injection ensures an efficient allocation of computational resources, balancing high-quality generation with reduced computational load.

To validate this approach, we first trained a model that incorporates the full set of control conditions, with $B$ layers and $T$ total diffusion steps. When all control conditions are included, the input sequence is represented as ${{h}^{total}}=[{{h}^{p}},{{h}^{z}},{{h}^{i}},{{h}^{s}},{{h}^{c}}]$, with the total sequence length given by ${{l}^{tot}}={{l}^{p}}+{{l}^{z}}+{{l}^{i}}+{{l}^{s}}+{{l}^{c}}$. At each timestep $t \in {1, ..., T}$ and layer $b \in {1, ..., B}$, the input to the multi-head attention mechanism consists of queries and keys, denoted as ${{Q}^{(b,t)}}$ and ${{K}^{(b,t)}} \in \mathbb{R}^{h \times {{l}^{tot}} \times d}$. We then measure the importance of the three visual input conditions through attention preference weights:
\begin{equation}\label{}
{{A}^{(b,t)}}=\mathbf{Softmax} (\frac{{{Q}^{(b,t)}},{{K}^{(b,t){\top}}}}{\sqrt{d}})\in {{\mathbb{R}}^{h\times {{l}^{tot}}\times {{l}^{tot}}}}
\end{equation}
For each control condition $ci$, we extract its corresponding subgraph, apply the relevant condition mask, and compute the full-dimensional mean to obtain a scalar value for each timestep and layer:
\begin{equation}\label{}
{{S}_{ci}}(b,t)=\mathbf{Mean}({{A}^{b,t,c}}\odot mas{{k}_{ci}})
\end{equation}
Here, the mask for the subject condition is applied to all areas outside the subject in the image, while the mask for the glyph condition covers regions outside the textual content.

Figure~\ref{fig-main-importance} illustrates the final visualization, showing the importance heatmaps of the three conditions across different timesteps and layer positions. From these maps, we observe that the background style dominates in the early layers and early timesteps but decays rapidly as the generation progresses. In contrast, the subject condition forms a persistent high-intensity band in the mid-to-deep layers, spanning most of the timesteps. The glyph condition, while exhibiting a lower overall magnitude, gradually increases in intensity in the middle layers and later timesteps, corresponding to the refinement of strokes and glyphs. Overall, the conditions display a non-uniform and complementary relationship across both timesteps and model depth. This analysis allows us to refine the selective injection strategy, ensuring that attention computations are retained for conditions with higher importance in ${{S}_{ci}}(b,t)$, while skipping less responsive locations, thus optimizing the overall efficiency of the model.

\section{Experiments}
\label{sec:experiments}

\subsection{Settings}

\begin{table*}[t]
\small
\centering
\caption{Quantitative results. The best scores are \textbf{bolded}, while the second-best is \underline{underlined}. Models marked with ``*" in the table indicate closed-source models, while ``-" denotes that the corresponding metric cannot be computed for the images generated by that model.}
\label{tab:quantitative}
\begin{tabular}{l|cc cc cc cc}
\toprule[1.1pt]
\multirow{2}{*}{Models}                      & \multicolumn{2}{c}{Visual Text Quality} & \multicolumn{2}{c}{Subject Consistency} & \multicolumn{2}{c}{Style Consistency} & \multicolumn{2}{c}{General Quality} \\ \cmidrule(lr){2-9}
                                             & Sen. Acc$\uparrow$                & NED$\uparrow$                & DINO$\uparrow$                & IoU$\uparrow$                & CSD$\uparrow$                & CLIP-I$\uparrow$            & IR-Score$\uparrow$          & FID$\downarrow$     \\ \midrule
Flux-inpaint~\cite{flux2024}                                 & -                   & -                  & 0.853               & 0.877              & -                  & -                 & 0.822             & 79.41   \\
Flux-Fill~\cite{flux2024}                                     & -                   & -                  & 0.867               & 0.932              & -                  & -                 & 0.950             & 75.12   \\ \midrule
Flux-Kontext~\cite{flux2024}                                 & -                   & -                  & 0.831               & 0.793              & 0.573              & 0.470             & 0.955             & 76.76   \\
OmniControl2~\cite{tan2025ominicontrol2}                                 & -                   & -                  & 0.812               & 0.685              & 0.692              & 0.477             & 0.930             & 80.13   \\
OmniGen2~\cite{wu2025omnigen2}                                     & -                   & -                  & 0.820               & 0.822              & 0.685              & 0.433             & 0.945             & 81.83   \\
USO~\cite{wu2025uso}                                          & -                   & -                  & 0.900               & 0.727              & 0.719              & 0.577             & 0.911             & 74.19   \\ \midrule
Glyph-ByT5-v2~\cite{liu2024glyphv2}                                & 0.685               & 0.815              & -                   & -                  & -                  & -                 & 0.837             & 82.63   \\
Anytext2~\cite{tuo2024anytext2}                                     & 0.721               & 0.582              & -                   & -                  & -                  & -                 & 0.701             & 86.35   \\ \midrule
PosterMaker~\cite{gao2025postermaker}                                  & 0.765               & 0.848              & 0.916               & 0.954              & -                  & -                 & 0.974             & 60.55   \\
Flux-Text \& inpaint ~\cite{flux2024}                         & 0.775               & 0.892              & 0.893               & 0.824              & -                  & -                 & 0.956             & 73.71   \\ \midrule
Qwen-Image-Edit ~\cite{wu2025qwen}                              & 0.831               & 0.960              & {\ul 0.922}         & 0.903              & 0.722              & \textbf{0.660}    & 0.994             & 69.86   \\ 
Seedream 4.0* ~\cite{seedream2025seedream}                                 & \textbf{0.865}      & {\ul 0.972}        & 0.864               & 0.837              & 0.700              & {\ul 0.639}       & {\ul 1.012}       & 64.21   \\
\textbf{Ours (Stage I)}           & {\ul 0.857}         & \textbf{0.976}     & \textbf{0.923}      & \textbf{0.972}     & \textbf{0.729}     & 0.582             & \textbf{1.036}    & \textbf{54.39}   \\
\textbf{Ours (Stage II)} & 0.847               & 0.969              & 0.914               & {\ul 0.960}        & {\ul 0.727}        & 0.594             & 0.995             & {\ul 55.24}   \\ \bottomrule
\end{tabular}
\vspace{-7pt}
\end{table*}

\noindent \textbf{Implementation Details.} 
Please see Appendix Sec.~\ref{appendix:implementation_details}.

\noindent \textbf{Comparative Methods.}
Since our approach is a multi-conditional generative method, we compare it not only with the open-source base model Flux~\cite{flux2024} and its variants, but also with state-of-the-art models across several relevant categories, including visual text generation, conditional image generation, and closed-source commercial models. Specifically, for visual text generation, we select Glyph-ByT5-v2~\cite{liu2024glyphv2} and AnyText2~\cite{tuo2024anytext2} as comparison baselines. For conditional image generation, we include Flux-Kontext~\cite{flux2024}, OminiControl2~\cite{tan2025ominicontrol2}, OminiGen2~\cite{wu2025omnigen2}, the subject-driven USO~\cite{wu2025uso}, and the poster generation model PosterMaker~\cite{gao2025postermaker}. In addition, we also incorporate an image generation foundation model Qwen-Image-Edit~\cite{wu2025qwen} and a closed-source commercial model   Seedream 4.0~\cite{seedream2025seedream}, as part of the comparison.

\noindent \textbf{Evaluation Metrics.}
Following previous works, we adopt multiple quantitative metrics to evaluate different aspects of our generated results, including visual text quality, subject consistency, background style consistency, and general image quality. Specifically, Sentence Accuracy (Sen. Acc) and Normalized Edit Distance (NED) are used to assess the accuracy of visual text generation within images. For subject consistency, we employ DINO score and IoU, where the subject regions are first extracted using Grounded SAM~\cite{ren2024groundedsam}, and the DINO score is computed as the cosine similarity between the generated and reference subject features in the DINO embedding space, while IoU measures their spatial overlap. To evaluate background style consistency, we use CSD \cite{somepalli2024csd} and CLIP-I, where CLIP-I represents the cosine similarity between generated and reference images in the CLIP embedding space, capturing global background similarity. Finally, IR-Score~\cite{xu2023imagereward} and FID are adopted to assess the overall perceptual fidelity and distributional quality of the generated images.


\subsection{Qualitative Analyses}

As shown in Fig.~\ref{fig:comp_gen}, the baseline models exhibit clear limitations when applied to the task of product poster generation. Flux-Fill and USO can only perform image synthesis driven by either a subject reference or both subject and background reference images, respectively, but neither is capable of generating images containing visual text, making them unsuitable for poster creation scenarios that require integrated textual elements. PosterMaker, while capable of generating images with both the specified subject and embedded text, struggles to maintain background style consistency, often failing to reproduce the desired stylistic characteristics of product posters.

In contrast, Qwen-Image-Edit and Seedream 4.0 demonstrate relatively good subject consistency and visual text rendering capabilities. Nonetheless, they often produce redundant or mismatched text, and their style transfer tends to exhibit a “copy-and-paste” effect, failing to generate diverse or contextually coherent backgrounds based on the style images. Our proposed InnoAds-Composer effectively overcomes these limitations, achieving strong consistency across text, subject, and background style. It is capable of generating visually appealing, semantically coherent, and stylistically controlled poster images, demonstrating its practicality for multi-conditional product advertisement generation.

\begin{figure*}[t]
\centering
\includegraphics[width=0.84\linewidth]{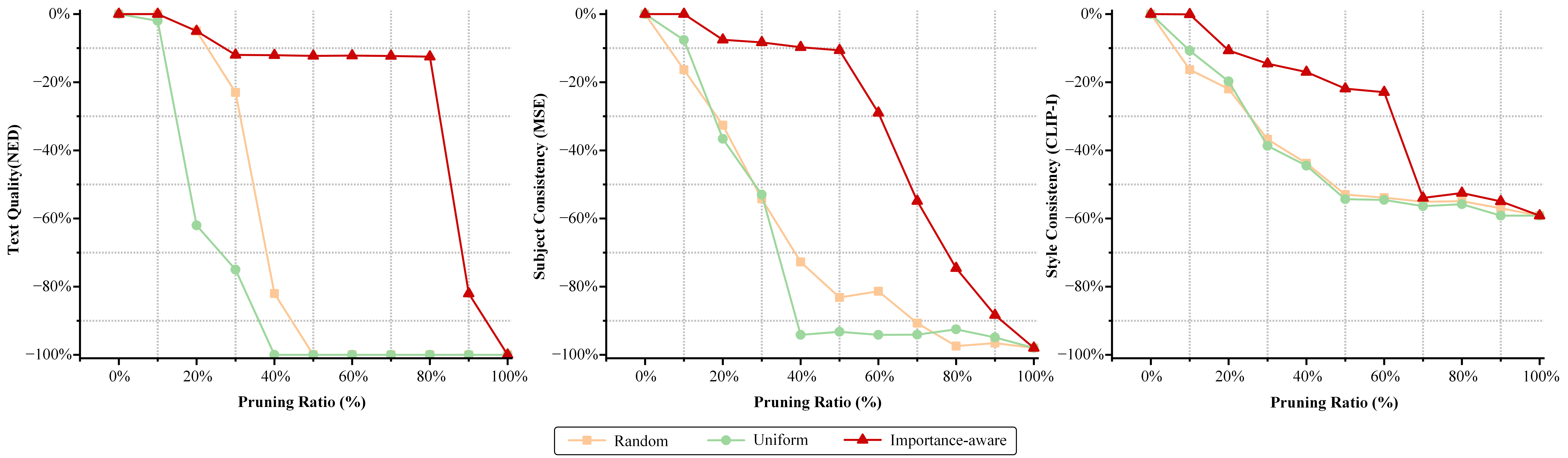}
\caption{Comparison of image generation quality under different condition token pruning strategies.}
\label{fig:ablation_importance}
\vspace{-8pt}
\end{figure*}

\subsection{Quantitative Evaluations}

Table~\ref{tab:quantitative} presents the quantitative comparison across all baselines and our proposed InnoAds-Composer under two configurations, Stage I and Stage II. Overall, in Stage I, our method achieves the best performance in nearly all aspects. It obtains the highest Sen. Acc of 0.857 and NED of 0.976, indicating superior visual text generation quality. In terms of subject consistency, it also leads with a DINO score of 0.923 and an IoU of 0.972, demonstrating strong alignment between the generated subjects and their references. For background style consistency, it attains a CSD of 0.729 and a CLIP-I score of 0.582, showing stable preservation of scene style. Moreover, it achieves the best overall image quality, reflected by an IR-Score of 1.036 and a FID of 54.39, significantly outperforming all open-source and commercial competitors. After Stage II, our method maintains comparable performance while improving computational efficiency. By performing importance analysis across layers and timesteps and routing each condition only to its most responsive regions, Stage II effectively shortens the active token sequence and curbs quadratic compute growth. Although it shows a slight decline in most metrics, its overall quality remains competitive, achieving a Sen. Acc of 0.847, DINO score of 0.914, and FID of 55.24. This balance between accuracy and efficiency highlights the scalability and practicality of the proposed multi-conditional generation framework.

\begin{table}[t]
\small
\centering
\caption{Efficiency analysis across different training stages.}
\label{tab:efficiency}
\resizebox{0.47\textwidth}{!}{
\begin{tabular}{c|ccc}
\toprule[1.1pt]
\textbf{Methods} & \textbf{Latency(s)} & \textbf{FLOPs(T)} & \textbf{Memery(G)} \\ \midrule
Flux-Kontext~\cite{flux2024}     & 76.02                  & 218.45            & 55.29              \\
Ours (Stage I)   & 55.87                  & 165.56            & 39.71              \\
Ours (Stage II)  & \textbf{47.32}                  & \textbf{135.25}            & \textbf{39.41}              \\ \bottomrule
\end{tabular}
}
\vspace{-10pt}
\end{table}

We further evaluate the computational efficiency of our method in terms of inference latency, FLOPs, and GPU memory consumption, as summarized in Table~\ref{tab:efficiency}. Leveraging decoupled attention, Stage I reduces latency by 26.5\% and FLOPs by 24.2\% compared to Flux-Kontext (which uses native full attention), with notably lower memory usage. Stage II builds on this by pruning redundant tokens and applying adaptive fine-tuning, achieving further reductions of 37.8\% in latency and 38.1\% in FLOPs—without compromising generation quality. These results demonstrate that our combined decoupled attention and token pruning strategy significantly improves efficiency while maintaining high-quality output, striking a strong balance between performance and resource usage.

\subsection{Ablation Study}

\begin{figure}[t]
\centering
\includegraphics[width=1\linewidth]{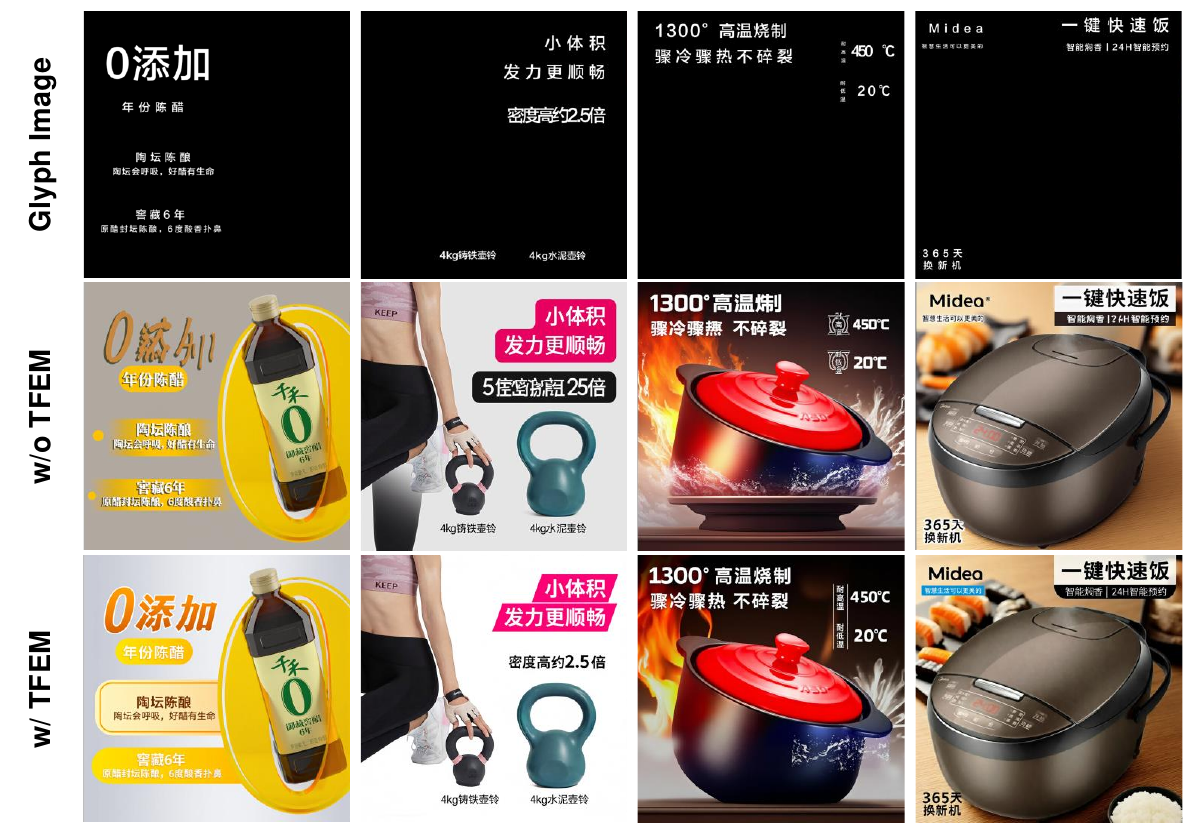}
\caption{Ablation study of Text Feature Enhancement Module.
Zoom in for details.
}
\label{fig:ablation_text_module}
\vspace{-10pt}
\end{figure}

\noindent \textbf{Effect of Text Feature Enhancement Module.}
As shown in Fig.~\ref{fig:ablation_text_module}, we conduct an ablation study on the Text Feature Enhancement Module. Without this module, the generated images exhibit noticeable textual errors, whereas incorporating it significantly improves the quality and accuracy of rendered text. Quantitatively, introducing the Text Feature Enhancement Module leads to an approximate 5\% improvement in Sen. Acc, demonstrating its effectiveness in enhancing visual text generation.

\noindent \textbf{Analyses of Importance-aware Condition Injection.} After the first-stage training, we evaluate three condition token pruning strategies during inference: random, uniform, and importance-aware. Generation quality is assessed using NED for text quality, MSE for subject consistency, and CLIP-I for style consistency. As shown in Fig.~\ref{fig:ablation_importance}, random and uniform pruning cause a rapid decline across all metrics, while the importance-aware strategy maintains stable quality until a large portion of uninformative tokens is removed. Specifically, glyph quality remains robust until about 80\% of tokens are pruned, and subject and style conditions preserve good performance up to roughly 50\% and 60\% pruning, respectively, before a sharp degradation occurs. Therefore, we adopt the importance-aware condition injection strategy and apply the corresponding token pruning ratios during the second-stage training to effectively reduce computational overhead.

\section{Conclusion}
\label{sec:conclusion}
We presented InnoAds-Composer, a single-stage, multi-condition framework for e-commerce poster generation that delivers simultaneously subject, text, and designed style. By unifying subject, style, and glyph (text) tokens in the same space and preserving the priors of an MM-DiT backbone, our approach enables joint inference without multi-stage pipelines. The proposed Text Feature Enhancement Module (TFEM) fuses single/entire glyphs with positional information, substantially improving text's sharpness, boundary integrity, and readability. Complementing this, we perform importance-aware condition injection and decoupled attention to reduce redundant interactions and shorten the activated sequence, leading to improved inference efficiency. To facilitate learning and fair comparison, we constructed InnoComposer-80K and the evaluation benchmark InnoComposer-Bench, covering subject, text, style, and overall quality. 
{
    \small
    \bibliographystyle{ieeenat_fullname}
    \bibliography{main}
}
\clearpage
\setcounter{page}{1}
\maketitlesupplementary

\setcounter{section}{0}
\section{Related Work}
\label{appendix:relatedworks}
\subsection{Poster Generation}
Poster generation aims to automatically produce visually appealing layouts that integrate images, text, and design elements to effectively convey information and aesthetic appeal. Recent advances such as COLE~\cite{jia2023cole}, Posta~\cite{chen2025posta} and PosterCraft~\cite{chen2025postercraft} leverage MLLMs to enable multi-stage control and iterative optimization, generating posters with high artistic quality and visual coherence. However, these methods are primarily designed for general or artistic compositions and are less suitable for visually appealing promotional images that must effectively present product information and attract consumer attention in e-commerce scenarios.

To address the specific requirements of e-commerce poster generation, several tailored approaches~\cite{zhao2025dreampainter,guo2025repainter,gao2025postermaker} have been proposed. DreamPainter~\cite{zhao2025dreampainter} and Repainter~\cite{guo2025repainter} introduce inpainting-based frameworks to customize both product and background regions, enabling controlled and coherent visual synthesis. PosterMaker~\cite{gao2025postermaker} further extends this line of work by combining prompt, subject, and text conditions to achieve fine-grained customization of background, product, and textual elements. Nonetheless, its reliance on prompt-based background generation often leads to results that deviate from desired visual or semantic constraints. 
\subsection{Text Rendering}
Text rendering aims to generate visually coherent and legible text within images, often requiring fine-grained control over font, layout, and contextual consistency. Early text rendering methods~\cite{chen2023textdiffuser,yang2023glyphcontrol,chen2024textdiffuser, zhu2024visual} primarily focused on generating Latin characters such as English text, but struggled to generalize to non-Latin scripts like Chinese due to the lack of corresponding text representations. To address this limitation, subsequent approaches introduced glyph-based representations~\cite{tuo2023anytext,tuo2024anytext2,ma2024chargen,jiang2025controltext,liu2024glyph,liu2024glyphv2} to bridge the gap between different languages. For instance, AnyText~\cite{tuo2023anytext} integrates glyph images as conditional inputs through a ControlNet~\cite{zhang2023adding} structure, enabling controllable rendering of multilingual text, while Glyph-ByT5~\cite{liu2024glyph} employs a customized multilingual text encoder trained on glyph representations to generate non-Latin characters effectively.

More recent studies~\cite{lan2025fluxtext,xie2025textflux,lu2025easytext,wang2025reptext,he2025plangen,
lu2025uni,bi2025customttt,ling2025mofu} have adopted Diffusion Transformer (DiT)~\cite{peebles2023scalable} architectures to achieve higher-quality and more contextually consistent text generation. TextFlux~\cite{xie2025textflux}, for example, uses Flux-Fill~\cite{flux2024} as its backbone and leverages in-context learning to better capture glyph structure and spatial dependencies. Building on this line of work, FluxText~\cite{lan2025fluxtext} further explores multiple condition fusion strategies to enhance the fidelity and controllability of generated text. Inspired by these advances, we adopt a DiT-based backbone in our framework to improve the quality, clarity, and contextual alignment of text generation in e-commerce poster synthesis.

\subsection{Multi-Condition Control Generation}

Controllable image generation aims to incorporate multiple conditioning signals—such as text, layout, or structural guidance—into the generative process to achieve fine-grained control over visual content. Earlier approaches typically relied on ControlNet~\cite{zhang2023adding} or IP-Adapter~\cite{ye2023ip} architectures to inject additional conditions through feature modulation or adapter networks. More recently, DiT-based methods~\cite{tan2025ominicontrol,zhang2025easycontrol,wu2025uno,wu2025qwen} have demonstrated strong potential for multi-condition control by integrating conditioning tokens directly into the denoising process. For example, OmniControl~\cite{tan2025ominicontrol} and UNO~\cite{wu2025uno} concatenate textual or semantic tokens with noisy image tokens to achieve unified conditional generation, while IC-Edit~\cite{zhang2025context} and insertanything~\cite{song2025insert} performs spatial concatenation and leverages in-context learning to support diverse conditional editing tasks.

However, as the number of conditions increases, the corresponding growth in token count leads to higher attention computation costs and reduced efficiency. To mitigate this issue, several studies~\cite{peng2024controlnext,ma2025efficient,anagnostidis2025flexidit,tan2025ominicontrol2,he2025fulldit2,ling2025ragar,wang2025adstereo,wang2024cost,zhang2025lgast} have explored more efficient conditioning mechanisms. OmniControl2~\cite{tan2025ominicontrol2}, for instance, computes condition token features only once and reuses them across denoising steps, while FullDiT2~\cite{he2025fulldit2} introduces a dynamic token selection mechanism to adaptively identify and retain the most informative context tokens during generation. Although these methods have achieved promising results in general visual synthesis tasks, applying multi-condition control to e-commerce poster generation remains challenging, as it requires simultaneously maintaining background style consistency, accurate text rendering, and product integrity while ensuring high-quality and efficient image generation.

\begin{figure*}[t]
\centering
\includegraphics[width=0.9\linewidth]{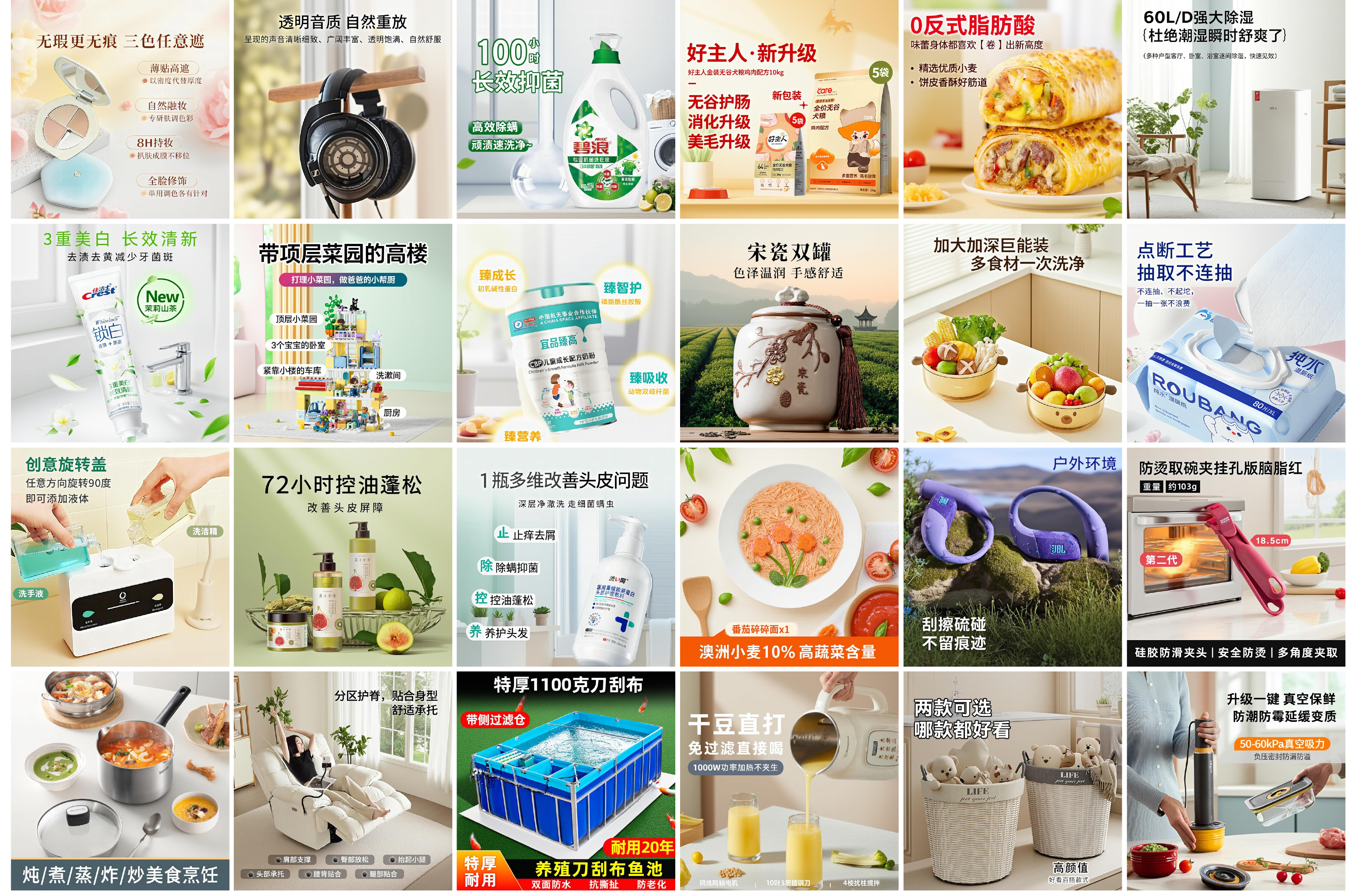}
\caption{Additional qualitative results generated by our method.
}
\label{fig:appendix_cases}
\vspace{-10pt}
\end{figure*} 

\section{Implementation Details}
\label{appendix:implementation_details}

InnoAds-Composer is developed based on the FLUX model~\cite{flux2024}, which is pretrained on a large-scale text rendering dataset  AutoPP1M~\cite{fan2025autopp} and possesses an intrinsic awareness of Chinese characters. The model is further optimized under tri-conditional control using our InnoComposer-80K dataset through a two-stage training strategy. (1) In \textit{Stage I}, we fine-tune all MM-DiT blocks using LoRA modules with a rank of 256. A constant learning rate of $2 \times 10 ^ {-5}$ is adopted, and the training process requires approximately 1.1k GPU hours. (2) In \textit{Stage II},  We removed selected tokens and fine-tuned the network to minimize performance degradation. During this stage, the learning rate was set to $1 \times 10 ^ {-6}$, and training was conducted for approximately 100 GPU hours based on the checkpoint from \textit{Stage I}. All training processes were conducted at a resolution of 800 x 800, using Ascend 910B. During the inference phase, to ensure a fair comparison with open-source models, we used the A100 for evaluating performance and inference latency. 
Besides, we have supplemented the pseudocode for TFEM as follows:
\begin{table}[t]
\centering
\renewcommand{\arraystretch}{1.1} 
\resizebox{0.45\textwidth}{!}{%
\begin{tabular}{l}
\hline
\textbf{Algorithm: Text Feature Enhancement Module (TFEM)}\\
\hline
\textbf{Input:} Glyph image $I_g$, Single-glyph crops $\{C_i\}_{i=1}^N$\\
\textbf{Output:} Enhanced glyph tokens $h^c$\\
\hline
1: $h^{c1} \leftarrow \mathrm{Patchify}(\mathrm{VAE\_Encode}(I_g))$ \hfill // Global structure branch\\
2: \textbf{for} each crop $C_i$ \textbf{in} $\{C_i\}_{i=1}^N$ \textbf{do}\\
3:\quad $f_i \leftarrow \mathrm{OCR\_Backbone}(C_i)$\\
4:\quad $p_i \leftarrow \mathrm{Add\_Positional\_Encodings}(f_i,\mathrm{abs\_pos},\mathrm{font\_size},\mathrm{local\_pos})$\\
5: \textbf{end for}\\
6: $h^{c2} \leftarrow \mathrm{Concat}(\{p_i\}_{i=1}^N)$ \hfill // Local semantic branch\\
7: // Character Encoder Fusion via Cross-Attention\\
8: $h^c \leftarrow \mathrm{Softmax}\!\left(\frac{(h^{c1}W_Q)(h^{c2}W_K)^{\mathsf T}}{\sqrt{d}}\right)(h^{c2}W_V) + h^{c1}$\\
9: \textbf{return} $\mathrm{LayerNorm}(\mathrm{FFN}(h^c))$\\
\hline
\end{tabular}%
}
\label{tab:tfem}
\end{table}



\begin{figure}[t]
\centering
\includegraphics[width=\linewidth]{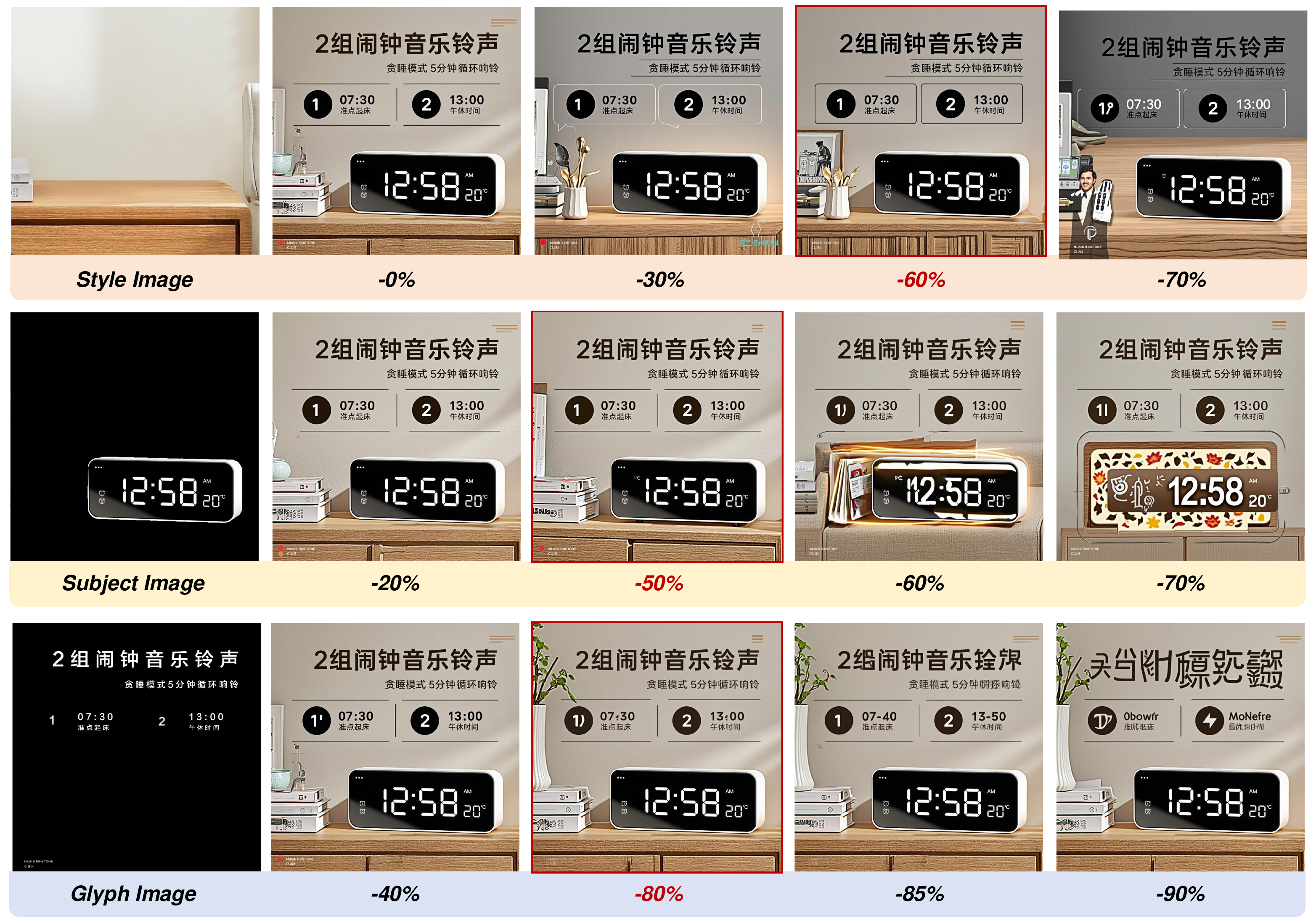}
\caption{Qualitative results under different condition token pruning ratios.}
\label{fig:appendix_stage_II}
\vspace{-10pt}
\end{figure}

\section{More Experiments}


\begin{table}[t]
\centering
\caption{Comparison of Stage II performance under different token pruning ratios. In the pruning ratio column, [x, y, z] denote the token pruning ratios for the glyph, subject, and style conditions, respectively.}
\label{tab:apendix_stage_II}
\begin{tabular}{cccc}
\toprule[1.1pt]
\textbf{Pruning Ratio ($\%$)} & \textbf{NED $\uparrow$} & \textbf{MSE $\downarrow$} & \textbf{CLIP-I $\uparrow$} \\ \midrule
{[}0,0,0{]}            & 0.976        & 0.056        & 0.582           \\
{[}50,20,30{]}          & 0.971        & 0.057        & 0.581           \\
{[}70,40,50{]}         & 0.970        & 0.057        & 0.562           \\
{[}80,50,60{]} (ours)  & 0.969      & 0.058        & 0.594           \\
{[}90,60,70{]}         & 0.582        & 0.066        & 0.451           \\ \bottomrule
\end{tabular}
\end{table}

\textbf{More Cases.}
\label{appendix:more_cases}
Fig.~\ref{fig:appendix_cases} presents additional generation results produced by our method. As shown, our approach not only maintains high-fidelity subject consistency for various products, but also delivers accurate and visually coherent text rendering. Moreover, the method produces realistic and diverse background styles, demonstrating its strong ability to integrate multiple conditions into cohesive, high-quality product posters.

\textbf{Different Token Pruning Ratios.} To validate the effectiveness of our importance-based token pruning ratios, we first evaluate alternative pruning proportions during \textit{Stage I} inference, with qualitative results shown in Fig.~\ref{fig:appendix_stage_II}. As illustrated, removing fewer tokens than the selected ratio preserves high-quality backgrounds, text rendering, and subject fidelity, while more aggressive pruning leads to a clear decline in generation quality. We further conduct \textit{Stage II} training under these alternative ratios, with quantitative results summarized in Table~\ref{tab:apendix_stage_II}. The table shows that pruning fewer tokens produces performance comparable to our chosen ratio, whereas pruning beyond it results in noticeable degradation across all metrics.

\end{document}